\documentclass{svproc}
\usepackage{url}
\usepackage{hyperref}
\hypersetup{breaklinks=true}
\usepackage{graphicx}
\usepackage{xcolor}
\usepackage{amsfonts}
\usepackage{amsmath}
\usepackage{comment}
\usepackage{ulem}

\begin{document}

\mainmatter              
\title{A New Lens on Homelessness: Daily Tent Monitoring with 311 Calls and Street Images}
\titlerunning{A New Lens on Homelessness}  
%

\author{Wooyong Jung\inst{1} \and Sola Kim\inst{2} \and
Dongwook Kim\inst{2} \and Maryam Tabar\inst{3} \and \\Dongwon Lee\inst{1}}

\authorrunning{Jung et al.} 

\institute{The Pennsylvania State University, USA\\
\email{wjung@psu.edu, dongwon@psu.edu}\\
\and
Arizona State University, USA\\
\email{sola@asu.edu, kim.dwt@asu.edu}
\and
University of Texas at San Antonio, USA\\
\email{maryam.tabar@utsa.edu}}

\maketitle              

\begin{abstract}
Homelessness in the United States has surged to levels unseen since the Great Depression. However, existing methods for monitoring it, such as point-in-time (PIT) counts, have limitations in terms of frequency, consistency, and spatial detail. This study proposes a new approach using publicly available, crowdsourced data, specifically 311 Service Calls and street-level imagery, to track and forecast homeless tent trends in San Francisco. Our predictive model captures fine-grained daily and neighborhood-level variations, uncovering patterns that traditional counts often overlook, such as rapid fluctuations during the COVID-19 pandemic and spatial shifts in tent locations over time. By providing more timely, localized, and cost-effective information, this approach serves as a valuable tool for guiding policy responses and evaluating interventions aimed at reducing unsheltered homelessness.
\end{abstract}

\section{Introduction}\label{sec1}

Homelessness in the United States has surged to its highest levels since the Great Depression, posing major challenges to public health, safety, and social equity. According to the 2024 Annual Homeless Assessment Report (AHAR), the number of people experiencing homelessness on a single night rose by 18.1\% from 2023, continuing an upward trend since 2016 \cite{hud2024}. While the Biden-Harris Administration’s American Rescue Plan (ARP) was the largest single-year effort to combat homelessness in U.S. history and slowed the increase between 2020 and 2022, the numbers have since rebounded as most ARP resources were exhausted \cite{hud2023}. Amid shifting federal priorities and a more rigid approach to street homelessness of the current administration, this increasing trend may continue for the time being \cite{cbs_homeless}.

To guide policy and resource allocation, the U.S. Department of Housing and Urban Development (HUD) mandates annual point-in-time (PIT) counts conducted by local Continuums of Care (CoCs) \cite{National_Alliance}. However, PIT counts face several critical limitations: they are infrequent (annual or biannual), geographically misaligned with standard Census boundaries, and methodologically inconsistent across CoCs \cite{tsai2022annual}. 
To address these issues, several improvement solutions have been suggested by government agencies and scholars. For instance, the Government Accountability Office (GAO) recommends enhancing standardized implementation guidelines, including data quality checks \cite{gao_2020}, which requires the involvement of data collection experts and relevant researchers \cite{tsai2022annual}. Some scholars also propose improvements to the longitudinal systems analysis and the introduction of post-enumeration and epidemiological surveys \cite{tsai2022annual}. However, implementing these surveys and collecting/maintaining the information is both costly and time-consuming.

\vspace{-.5mm}

Other studies have attempted to enhance the PIT data by mapping CoC-level counts to Census geographies using demographic or density-based disaggregation \cite{almquist2020connecting,byrne2013new}.
Although this approach improves spatial resolution, it introduces noise and still lacks the temporal granularity needed to monitor seasonal or short-term trends. Additionally, major U.S. cities such as New York City and San Francisco have been counting their homeless populations quarterly to supplement the annual PIT count. However, this method is still susceptible to measurement bias due to the reliance on single-day counts in each quarter. 
Furthermore, conducting these counts every three months is quite time-consuming and resource-intensive. With recent advancements in computing capacity, many scholarly trials have utilized sophisticated computational methods to tackle housing and neighborhood issues\cite{hwang2023systematic,jung2024hotspots,tabar2022forecasting}. However, to our knowledge, there has been limited research focusing on the use of computational techniques or new data sources to monitor trends in homelessness.

\vspace{-.5mm}

Concerning these challenges and in response to the necessity of alternative or supplementary data sources that can offer finer-grained insights into homelessness trends, this study explores the use of publicly available, crowdsourced data---specifically 311 Service Calls and street-level imagery---to track and predict the presence of homeless tents at high spatial (0.1 mile by 0.1 mile bounding box) and temporal (daily) resolution. We focus our analysis on San Francisco, a city where homelessness has continued to rise, with a 7\% increase in the homeless population since 2022 \cite{sfpit2024}. By integrating 311 call records (true positive reports) with geolocated street-view imagery (true positives and negatives) in the city, we aim to answer the central research question:
\textbf{\emph{Can crowdsourced data improve the spatial and temporal accuracy of homelessness monitoring beyond existing PIT-based methods?}}

\vspace{-.5mm}

Our results demonstrates that the combination of 311 Service Calls and street-level imagery provides significantly improved spatial and temporal granularity in monitoring homeless tent trends compared to traditional PIT counts. Our model successfully identified key city-level trends, such as the spike during the COVID-19 pandemic. Additionally, the model revealed a notable shift in spatial distribution over the past decade, with tent locations expanding beyond downtown area. These findings can guide more precise and timely policy interventions and facilitate better evaluation of homelessness initiatives. 

\section{Data}
Our two main data sources are the 311 Service Calls records and Mapillary street-view sequenced images. The 311 Service Calls dataset contains call record data from the 311 call centers across the U.S. While the types of service requests can vary depending on local governments, the dataset typically includes several hundred request types that impact the quality of neighborhoods and communities such as noise issue, abandoned vehicles, dead animal pickup, blocked street etc.\cite{dallasopendataServiceRequests}. In this study, we focus on service requests related to homeless tents. Mapillary is a crowdsourced map platform where users upload street-level images, which are accessible through its API. 

The 311 Service Call records and Mapillary images provide important temporal and spatial information about homeless tents, including the dates of reports and image captures, and their locations. The 311 records are mostly verified true positives. This means that when a specific location is reported for the presence of homeless tents, and an outreach team confirms it, we can be confident that at least one tent is present at that location at that specific time. However, not all tents are reported, as some areas may be more tolerant or some tents may not be causing enough of a disturbance for people to feel compelled to report them. Consequently, the 311 Service Call data does not allow us to identify true negatives throughout the city, and we cannot assume that there are no homeless tents in areas where no reports have been made. To fill this gap, we used street-level images from the Mapillary API, which consist of actual photographs of street scenes and cover almost every inch of the city at multiple times.

\begin{figure}[tb]
\includegraphics[width=\textwidth]{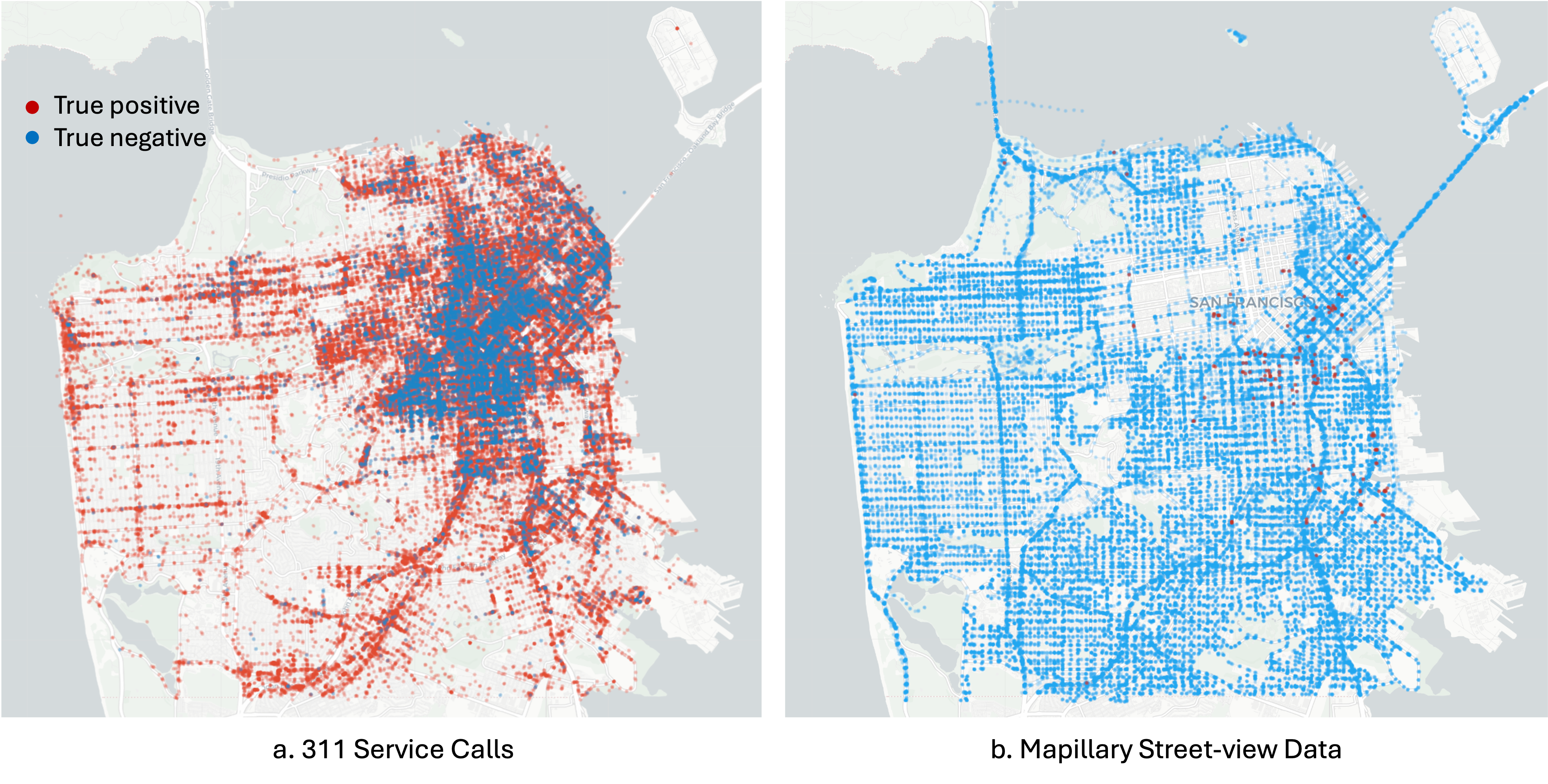}
\caption{(a) Locations reported for homeless tents through the 311 Service Calls \\ (b) Locations captured in the Mapillary street-level images throughout San Francisco} 
\label{fig_data_map}
\vspace{-3mm}
\end{figure}

Figure \ref{fig_data_map} illustrates the locations reported for homeless tents through the 311 calls on the left  and the locations where street-view images were captured using the Mapillary platform on the right, all within the San Francisco area from January 2016 to May 2024. The data points from the 311 calls are primarily concentrated in downtown areas and along specific streets, whereas the Mapillary data is more evenly distributed throughout the city. In the figure, red points represent spots where tents are present (true positive cases), while blue points indicate spots without tents (true negatives). For the 311 call dataset, true negative cases refer to instances where a location was reported for homeless tents, but the outreach team found no tents at that site. 
This means we can confidently categorize those locations as true negative, as no tents were present at that specific spot and time.

Along with the two primary datasets, we also used daily weather data, including daytime highs, nighttime lows, and precipitation obtained from Weather Underground\footnote{\url{https://www.wunderground.com/history/daily/us/ca/san-francisco/KJMC}}, as well as demographic information at the Census Block Group (CBG) level from the American Community Survey (ACS), such as population size, median household income, and the proportions of Black and White populations, to improve our model's predictive performance.

A more detailed explanation of how we extracted images and identified homeless tents is provided in the following Methods section. The final data points used in our model consist of 228,183 true positive cases and 43,809 true negatives after removing overlapping data points.

\vspace{-4mm}
\section{Methods}
\vspace{-2mm}
We followed a structured process that involved two key steps: (i) developing a machine-learning model to detect homeless tents from street-view images and (ii) constructing a trend-monitoring and forecasting model with the combined data of the tent information from the first step and the 311 call records. In the following subsections, we will go through the details of each step.
\vspace{-3mm}

\begin{figure}[h!]
  \centering
  \includegraphics[width=.95\textwidth]{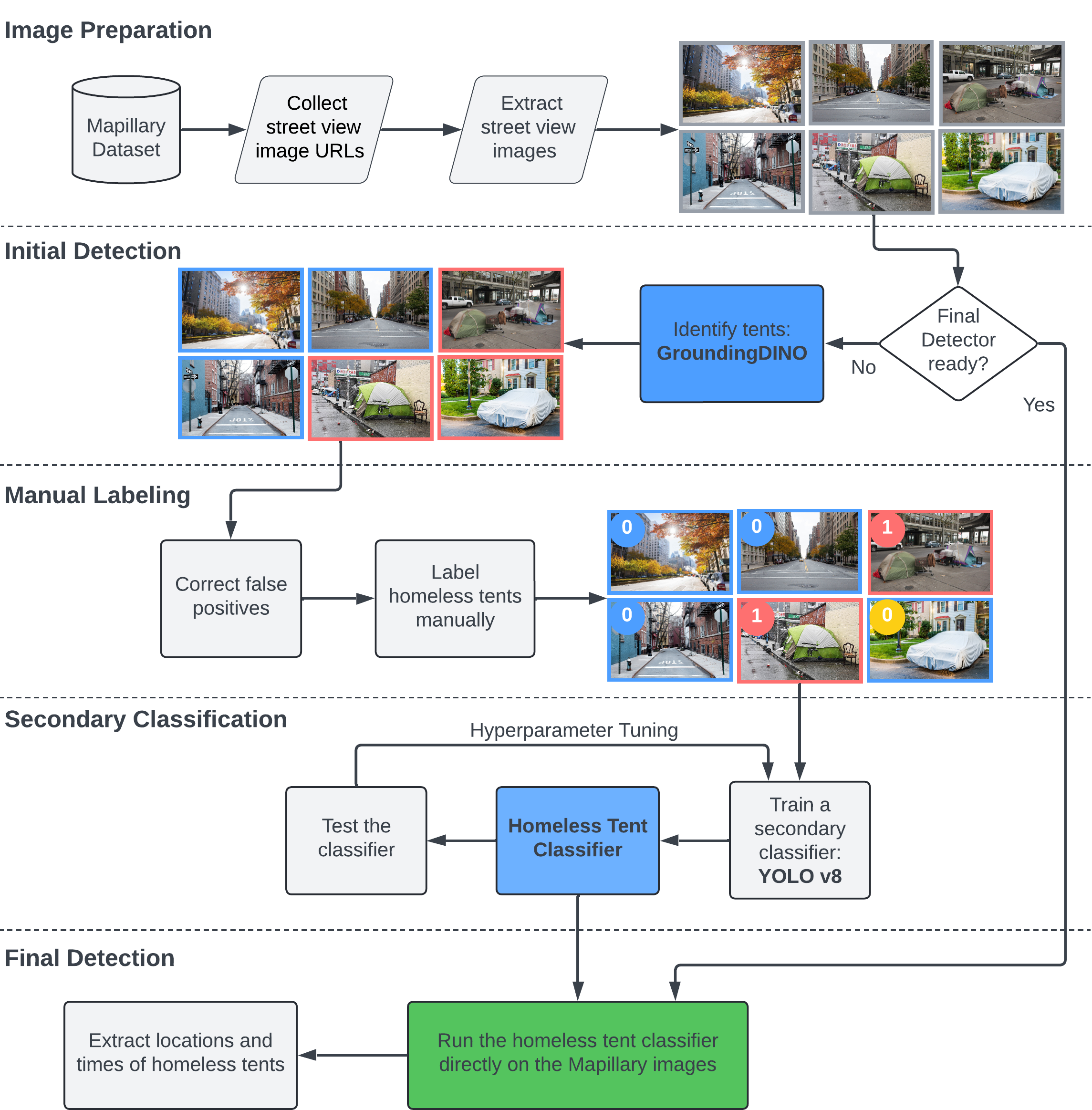}
  \caption{Homeless tent identification process} 
  \label{fig_method_framework}
  \vspace{-5mm}
\end{figure}

\subsection{Homeless Tent Detector} 
\vspace{-1mm}
Figure \ref{fig_method_framework} outlines the process of identifying homeless tents. We collected street-view images taken in San Francisco between January 2016 and May 2024 using the Mapillary API and applied GroundingDINO, a zero-shot object detection model that does not require prior training on labeled examples of homeless tents \cite{liu2023grounding}. This saved considerable time and effort in building a training set. After initial detection, we manually reviewed positive results to filter out false positives, such as vehicles with tarps or commercial tents, which GroundingDINO often misclassified. 

To minimize Type II errors (i.e., missed true positives), we adopted a lenient detection threshold of 0.4 in GroundingDINO\footnote{GroundingDINO applies \textit{box} and \textit{text} thresholds to filter detections; lower thresholds reduce false negatives but increase false positives \cite{liu2023grounding}.}. While this increased false positives, it ensured we captured more true tent instances. Also, the zero-shot detection model remarkably reduced the manual workload: from approximately 1.9 million collected images, only about 2,000 required review, achieving a reduction ratio of nearly 1:1,000.


After manually correcting false positives, we trained a secondary classifier, \textit{YOLOv8l} \cite{yolov8_ultralytics}, on the verified images. This secondary classifier aims to replace the initial detection and manual correction process for the images collected from the Mapillary database. We successfully trained the model, tuned the hyperparameters, and finalized the secondary classifier. Then, instead of going through the GroundingDINO and manual correction steps, we directly send the images obtained from the Mapillary API to the secondary classifier.


Once we extracted spatial and temporal tent data, we merged it with 311 Service Call records based on time and location. To avoid duplicate counts, we excluded entries if they were on the same day and within a 10-meter diameter of each other.
\vspace{-3mm}

\subsection{Trend Monitoring and Forecasting Model}
\vspace{-1mm}
To efficiently capture the spatial and temporal characteristics of our data, we aggregated it using space-time cubes—three-dimensional units combining spatial (x, y) and temporal (z) dimensions \cite{arcgisDescribeSpace}. Each spatial unit is a 0.1-mile by 0.1-mile square, and each temporal unit spans one day, from 00:00:00 AM to 11:59:59 PM. This approach helps reduce biases from missing data and measurements while compromising some spatial and temporal granularity. The final dataset consists of 13,461,046 space-time cubes, covering 4,379 spatial bounding boxes covering the city of San Francisco across 3,074 days (January 1, 2016 to May 31, 2024).

Using this dataset, we implemented a spatiotemporal variational Gaussian Process (ST-VGP) model to track and forecast trends in homeless tent presence. ST-VGP extends Gaussian Processes (GPs) to spatiotemporal settings while addressing scalability issues \cite{hamelijnck2021spatio}. Its probabilistic nature allows for quantifying predictive uncertainty, which is valuable for proactive policymaking \cite{roberts2013gaussian}. Unlike traditional time-series models, which struggle with irregular observations, ST-VGP handles sparse, unevenly spaced data effectively—making it well-suited for modeling the rare and irregular patterns of homeless tent appearances.
\vspace{-5mm}

\subsubsection{Model Formulation}
Let $s = 1,2,\dots,4379$ denote the spatial bounding boxes and $t = 1,2,\dots,3074$ represent the daily observation over 3,074 days. Then $Y_{s,t} \in \mathbb{N}_0$ is the observed homeless tent counts at location $s$ and on day $t$. Also, let the vector of covariates denote $X_{s,t} = [$\textit{precipitation}$_{s,t}$, \textit{daytime highest temperature}$_{t}$, \textit{nighttime lowest temperature}$_{t}$, \textit{white population}$_{s,t}$, \textit{black population}$_{s,t}$, \textit{household income}$_{s,t}]$\footnote{Due to the unavailability of bounding box-level daily demographic information, we utilize census block group-level data from American Community Survey (ACS) for the black and white populations and household income.}.

Considering the modest range of observed tent counts (minimum: 0, maximum: 15) and limited variation, Poisson likelihood function was used, such as $Y_{s,t}|f_{s,t} \sim \text{Poisson}(\mu_{s,t})$ with $\mu_{s,t} = \exp(f_{s,t})$, where $f_{s,t}$ is latent function. The latent function $f$ follows a Gaussian Process (GP) such that $$f\sim GP(m, \kappa((s,t,X),(s',t',X'))$$ where $m$ is a constant mean of $\log y_{s,t,X}$ and $\kappa(\cdot,\cdot)$ is a kernel function. To capture the more expressive interaction and similarity between the spatial, temporal, and other covariate variations, we used a hybrid kernel with both addition and multiplication such that $\kappa = \kappa_s + \kappa_t + \kappa_X + \kappa_s\times\kappa_t\times\kappa_X$, where $\kappa_s$, $\kappa_t$, and $\kappa_X$ represent spatial, temporal, and covariate kernel, respectively. For the spatial and temporal kernels, the Matérn kernel was used, with which the temporal aspect of the GP prior can be mapped to a state-space representation for the temporal process at each spatial location. This enables us to perform fast inference using Kalman filtering and smoothing implicitly \cite{hamelijnck2021spatio}. For the covariate kernel, the stationary RBF kernel was selected.


To address the scalability issue and reduce computational burden, we used 700 inducing points ($Z$) for the spatial aspect such that $\{Z_s\}_{s=1}^{700}$ and variational inference to incorporate the Poisson likelihood. The 700 points consist of the top 400 homeless tent hotspots and 300 random spots. Then, approximate posterior becomes $q(f,u)=p(f|u)q(u)$ with variational distribution $q(u)=N(m_u,S_u)$, where $u=f(Z_s),\: Z_s\in\mathbb{R}^{700\times2}$. We get inference done by maximizing the variational evidence lower bound (ELBO).
\vspace{-5mm}

\subsubsection{Model Calibration}
During model training, we utilized spatiotemporal cross-validation for parameter tuning. In this approach, we hold out the union of the time block and spatial block sequentially and train on their intersection, making sure that the model is tested with both new regions and future time points simultaneously. We compared the root-mean-square error (RMSE) for point predictions and negative log predictive density (NLPD) for uncertainty fit across various setups of the number of inducing points, output and length scales of the kernel functions, sample size, and learning rate. Since our goal was to evaluate whether the crowdsourced data is effective for monitoring the homeless tent trends, rather than to make precise count predictions, we prioritized setups with the lowest NLPD first. Then, we compared the RMSEs of these selected setups and finally chose the one with the lowest RMSE.

Once we predict the bounding-box-level tent counts, we aggregate these counts to determine city-level totals. The city-level data is the only available PIT ground-truth data from the San Francisco local government. We use a Monte Carlo (MC) simulation of aggregation based on the identified posterior distributions of each bounding box. Instead of relying on the raw expected rate, $\lambda$, we set a threshold, $\theta$, to prevent inflation that can occur from summing thousands of small expected values. Therefore, $P(Y_i>0)=1-\exp{(-\lambda_i)} < \theta$, only including bounding boxes with at least a $100\times\theta$\% chance of one or more tents. By calculating and comparing the RMSE between the ground-truth quarterly tent counts and our predicted city-level totals, we chose $\theta$ to be 0.7.
\vspace{-5mm}
\section{Results}
\vspace{-3mm}
\subsection{City-Level Trends}
\begin{figure}[t]
  \centering
  \includegraphics[width=.95\textwidth]{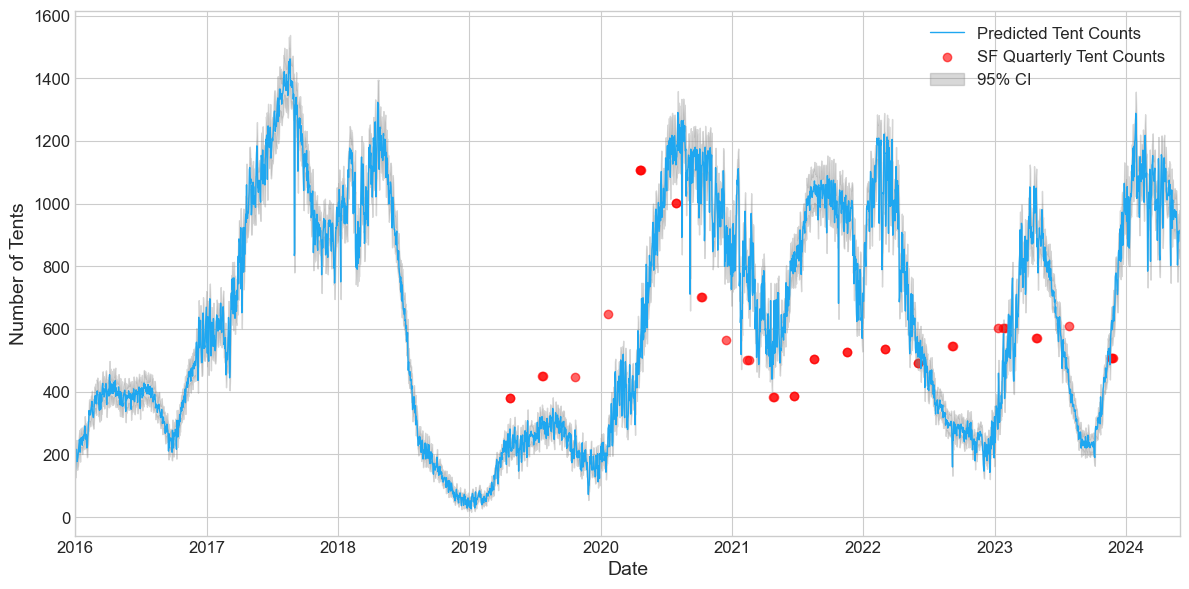}
  \caption{Predicted daily city-wide homeless tent counts}
  \label{fig_prediction}
  \vspace{-4mm}
\end{figure}

\vspace{-1mm}
Figure \ref{fig_prediction} illustrates the predicted trends of homeless tents, as marked by the blue line in San Francisco, from January 2016 to May 2024. The red dots represent the quarterly counts of tents by the local government.\footnote{The specific dates when tents were counted each quarter are not publicly available. We requested this information from the Department of Emergency Management (DEM) of San Francisco.} Since, for most quarters, volunteers count tents in different areas over two or three days, either sequentially or sporadically within the quarter, and then sum the counts, there is some uncertainty in the ground-truth data. Additionally, because it is unclear which areas are counted on which dates, accurately comparing the predicted counts with the actual counts is difficult. Despite these limitations, our model captures similar patterns to some extent, such as the rising trend since early 2020, followed by a decline until April 2021, and then a rebound. However, since 2022, our model predicts another spike, but the quarterly tent counts show a relatively modest increase. Another advantage of using our model is that it can predict the trends from 2016 to 2018, even when the government did not start counting tents quarterly.

One important aspect to note is the high volatility of the daily tent counts in our dataset. Especially from January 2020 to June 2022, when COVID-19 began to spread rapidly, the rolling standard deviation with a 7-day window is about 73, indicating that weekly tent counts increased or decreased by an average of 73 in this period. This shows that quarterly tent counts, which are the total of a few single-day counts, can be misleading about overall trends, and that, in turn, could give policymakers the wrong signal. 
\vspace{-5mm}

\subsection{Bounding Box-Level Counts}
\begin{figure}[t]
  \centering
  \includegraphics[width=.95\textwidth]{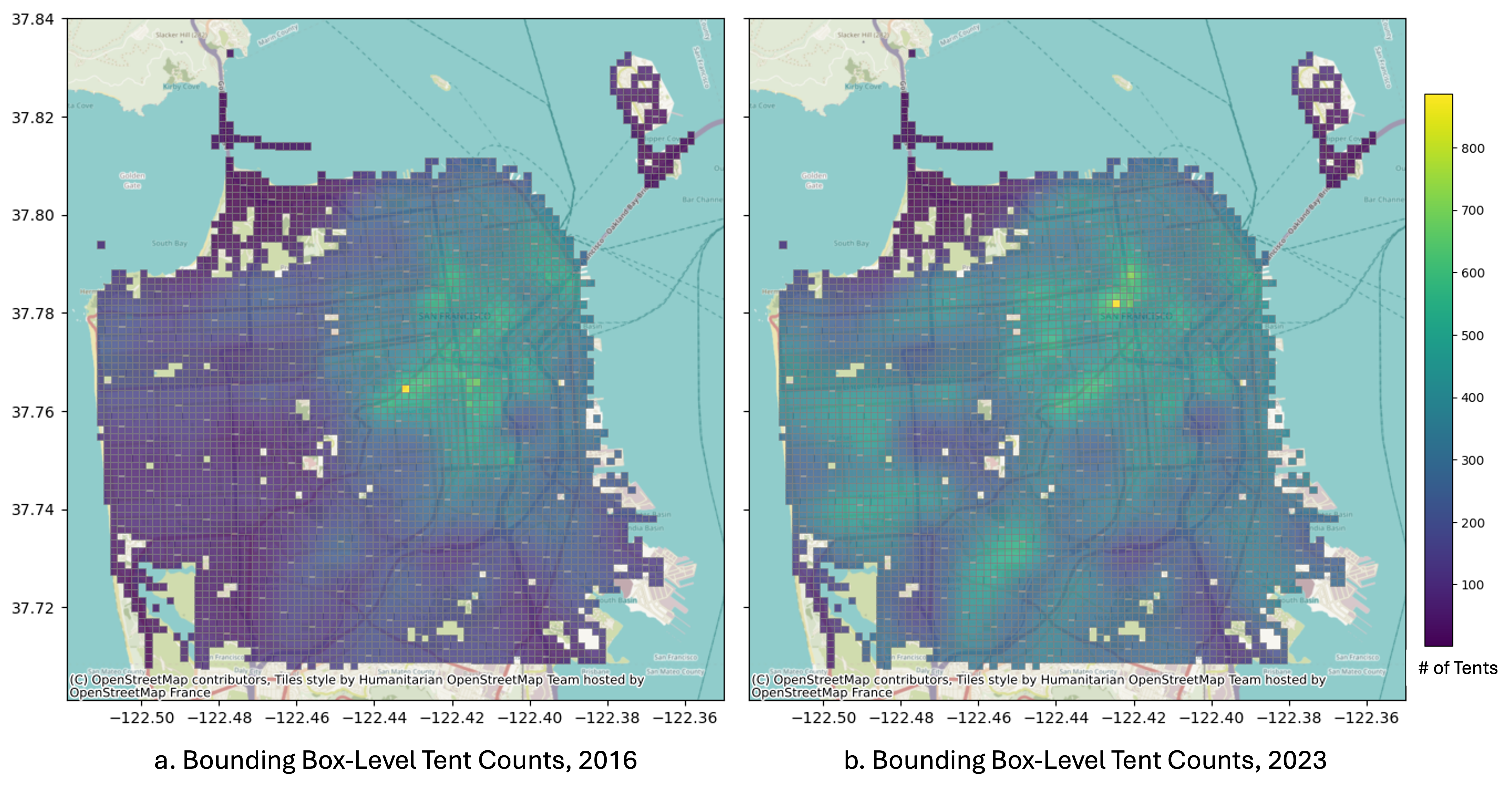}
  \caption{Bounding box-level homeless tent counts (a) in 2016 and (b) in 2023}
  \label{fig_bbox_heatmap}
  \vspace{-5mm}
\end{figure}

\vspace{-1mm}
In addition to the finer granularity in the temporal dimension as shown above, our model provides higher resolution in the spatial dimension. As demonstrated in Figure \ref{fig_bbox_heatmap}, the distribution of homeless tents has shifted since 2016. Back in 2016, most tents were concentrated in downtown and the Mission District. However, by 2023, this trend has expanded to various areas of the city, including the Sunset District.

Compared to the existing PIT counts, which provide homeless people information at the CoC or state level\footnote{\url{https://www.hudexchange.info/programs/coc/coc-homeless-populations-and-}\\\url{subpopulations-reports/}}, this high-resolution spatial information from our model can be usefully utilized. This allows policymakers and practitioners to direct outreach and intervention program precisely and supports more robust policy evaluation. 
\vspace{-5mm}

\section{Conclusion and Discussion}
\vspace{-3mm}
This study provides strong evidence of the usefulness of new data sources---311 Service Calls and street-view imagery---to monitor daily trends in homeless tent locations. This innovative use of the data sources improves the existing annual-based PIT counts and some local governments' quarter-based tent counts in terms of not only spatial and temporal granularity but also cost efficiency and resource optimization with the finer-grained information.

Since the two data sources are updated daily, our method can provide almost real-time homeless tent information in a city. Using this approach alongside PIT data can support more timely and informed decisions, enabling more efficient allocation of outreach teams, sanitation, and healthcare services.


Moreover, our trend monitoring model provides a valuable tool for evaluating policy impacts. For instance, to evaluate how the ``Housing First" approach impacts the number of homeless tents on the streets in the short term and the long term, policymakers and practitioners currently have to wait for the PIT counts, or quarterly tent counts, to be released after the intervention is implemented. Even with the counts, due to the daily variations resulting from other factors, such as weather conditions, it can be challenging to isolate the sole impact of the policy. With our model, they can track the daily change in the number of tents, along with the uncertainty level, after their intervention. This would greatly reduce the potential inaccuracy of the infrequent tent counts and be helpful in monitoring the trends. Another advantage of our approach, which incorporates the strengths of the Bayesian approach, is that as more data accumulates, our model evolves more accurately and delicately by updating the GP prior. 

Despite these advantages, this study has several limitations. First, our model focuses only on homeless individuals living in tents, excluding other types of homeless residents, such as those living in RVs. Additionally, while we emphasize the usefulness of the crowdsourced data, the coverage of this data is still inconsistent across different regions. Generally, areas with a larger population tend to have better quality data available. In future studies, addressing and improving these shortcomings will be valuable contributions.
\vspace{-2mm}

\vspace{-2mm}
\bibliography{sn-bibliography}
\bibliographystyle{splncs03}

\end{document}